%% file: RCSD.tex
\documentclass{article}
\usepackage{amsmath, amssymb}
\usepackage{algorithm}
\usepackage{algcompatible}
\usepackage{float}

 \usepackage{algcompatible}
 
\usepackage{xspace}

\usepackage{PRIMEarxiv}

\usepackage[T1]{fontenc}    
\usepackage{hyperref}       
\usepackage{url}            
\usepackage{booktabs}       
\usepackage{amsfonts}       
\usepackage{nicefrac}       
\usepackage{microtype}      
\usepackage{lipsum}
\usepackage{fancyhdr}       
\usepackage{graphicx}       
\graphicspath{{figure/}}     
\usepackage{makecell}
\usepackage{bbm}
\usepackage{algpseudocode}

\usepackage{amsmath}
\usepackage{subcaption}
\usepackage{cleveref} 
\usepackage{multirow}

\usepackage{hyperref}
\hypersetup{
    colorlinks=true,
    linkcolor=blue,
    citecolor=blue,
    urlcolor=orange
}
\usepackage{algpseudocode}
\usepackage{mathtools}

\usepackage{xcolor}
\title{	RCSD: Reinforcement and Continual Learning for Domain-Specific Speculative Decoding}

\author{
  Mingkang Liu (\texttt{liumingkang@hisense.com}), 
  Weiming Meng,
  PeiPei Jiao, \\
  Jiawei Zhang,
  Tianyuan Liu,
  Junyan Li,
  Lichen Pang (\texttt{cliff.pan@hisense.com}) \thanks{code can be released at https://github.com/ooolmk/RCSD-reinforcement-continue-learning-for-speculative-sampling},
}

\begin{document}

\maketitle

\begin{abstract}

Speculative decoding accelerates large language model inference by employing a compact draft model to propose candidate tokens that are verified in parallel by a target model. Conventional training relies on soft-label distillation, which forces the draft model to mimic the target's full output distribution—including high-entropy regions that degrade acceptance rates and increase early rejection risk. We propose RCSD (Reinforcement and Continual Learning for Speculative Decoding), a training framework that aligns draft model optimization with the system-level objective of maximizing acceptance length. RCSD leverages naturally occurring accepted and rejected sequences during speculative inference to construct implicit preference pairs, with masked loss to avoid penalizing target-aligned tokens. Evaluated across multiple benchmarks, RCSD improves average acceptance length by 35\% over HASS and achieves 1.35× faster end-to-end inference than HASS on practical workloads, without compromising generation quality.

\end{abstract}

\input{introduction}

\input{related-work}

\input{methodology}

\input{experiments}

\input{conclusion}

\bibliographystyle{abbrv}  
\bibliography{references}

\end{document}

%% file: introduction.tex
\section{Introduction}

\label{sec:introduction}

Large language models (LLMs) have demonstrated remarkable capabilities across a wide range of tasks. However, their autoregressive nature leads to high computational costs and significant latency during inference, which hinders their deployment in real-time applications. To address this bottleneck, speculative decoding has emerged as a prominent acceleration paradigm\cite{specdecode, medusa}. This approach leverages a fast draft model to generate multiple candidate tokens speculatively, which are then verified in parallel by the larger, more accurate target model. By accepting or rejecting these candidates in bulk, speculative decoding can substantially improve the overall throughput of the system.

\begin{figure}[h!]
\centering
\includegraphics[width=0.5\linewidth]{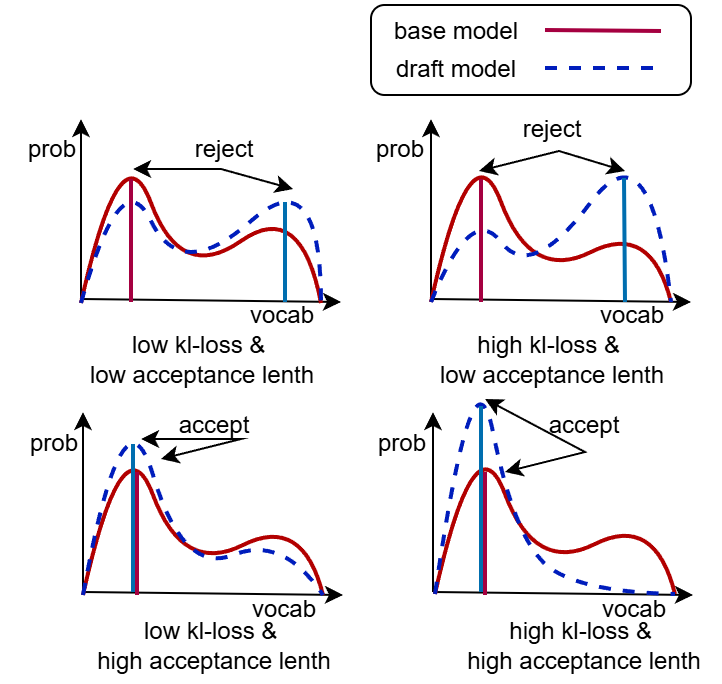}
\caption{Misalignment between KL divergence and acceptance rate. Due to limited parameters, the draft model cannot perfectly replicate the target distribution. However, low KL loss may still lead to token rejection. This discrepancy arises because speculative decoding acceptance depends on the relative ranking of top candidate tokens rather than exact distribution matching.}
\label{fig:acclen_klloss}
\end{figure}

The prevailing training strategy for draft models relies heavily on soft-label distillation\cite{eagle}, where the draft model is supervised to mimic the full output probability distribution of the target model. This method implicitly minimizes the forward Kullback-Leibler (KL) divergence, compelling the draft model to cover all modes of the teacher's distribution, including high-entropy regions with low-probability tokens. While it is theoretically true that minimizing the KL divergence to zero would result in the draft model perfectly replicating the target model and thus achieving the maximum possible acceptance length, this goal is practically unattainable. The draft model, being orders of magnitude smaller than the target, lacks the capacity to faithfully reproduce the complex distribution of its teacher. However, exact probability replication is not requisite. To maximize acceptance length, it suffices for the relative ordering of the top-k most probable tokens in the draft model to align with that of the base model. Consequently, a fundamental misalignment arises: the training objective (distribution matching) diverges from the ultimate system-level goal (maximizing the number of accepted tokens). For instance \Cref{fig:acclen_klloss}, a draft model might accurately capture the relative ranking of the top-$k$ tokens but fail to precisely replicate their exact probabilities, leading to a perfect acceptance rate despite a large distillation loss. 

This misalignment motivates our core insight: the draft model should transcend the role of a faithful imitator to become an efficiency-oriented predictor. Specifically, we advocate for a ‘mode-seeking' behavior rather than ‘mode-covering’, thereby concentrating predictive capacity on sequences with high acceptance probability, even if this necessitates disregarding low-probability tails of the target distribution. Ultimately, the optimization goal is redefined: rather than minimizing the divergence of logits, the system aims to maximize the expected acceptance length.

To bridge this gap, we propose \textbf{RCSD (Reinforcement \& Continual Learning for Speculative Decoding)}, a novel training framework that directly aligns the draft model's optimization with the system's performance metric. RCSD capitalizes on the intrinsic feedback loop of speculative decoding, where every verification step naturally produces pairs of accepted and rejected token sequences. We leverage these implicit preference signals to train the draft model using reinforcement learning principles. Specifically, we adapt two powerful algorithms: \textbf{Direct Preference Optimization (DPO)}\cite{DPO} and \textbf{Group-wise Relative Policy Optimization (GRPO)}\cite{GRPO}. In our DPO formulation, a sequence sampled purely from the target model serves as the ``chosen'' response, while a sequence generated through periodic alternation between the target and draft models (every $k$ steps) acts as the ``rejected'' response. Our GRPO approach constructs groups of sequences with varying drafting strategies (different $k$-step intervals) and uses the acceptance rate over these sequences as a direct and verifiable reward signal to optimize the expected acceptance length. 

Our experiments on both Qwen3-1.7B and Llama-3.1-8B demonstrate the efficacy of RCSD. It achieves an average improvement of 35\% in acceptance length over conventional distillation baselines(HASS) \cite{HASS} and delivers a 1.35$\times$ speedup on practical workloads like ShareGPT, all without compromising the quality of the final output. To our knowledge, this work is the first to systematically integrate preference optimization techniques like DPO and GRPO into the training pipeline for speculative decoding, establishing a new paradigm for building highly efficient draft models.


\begin{figure}[t]
  \centering
  \includegraphics[width=\linewidth]{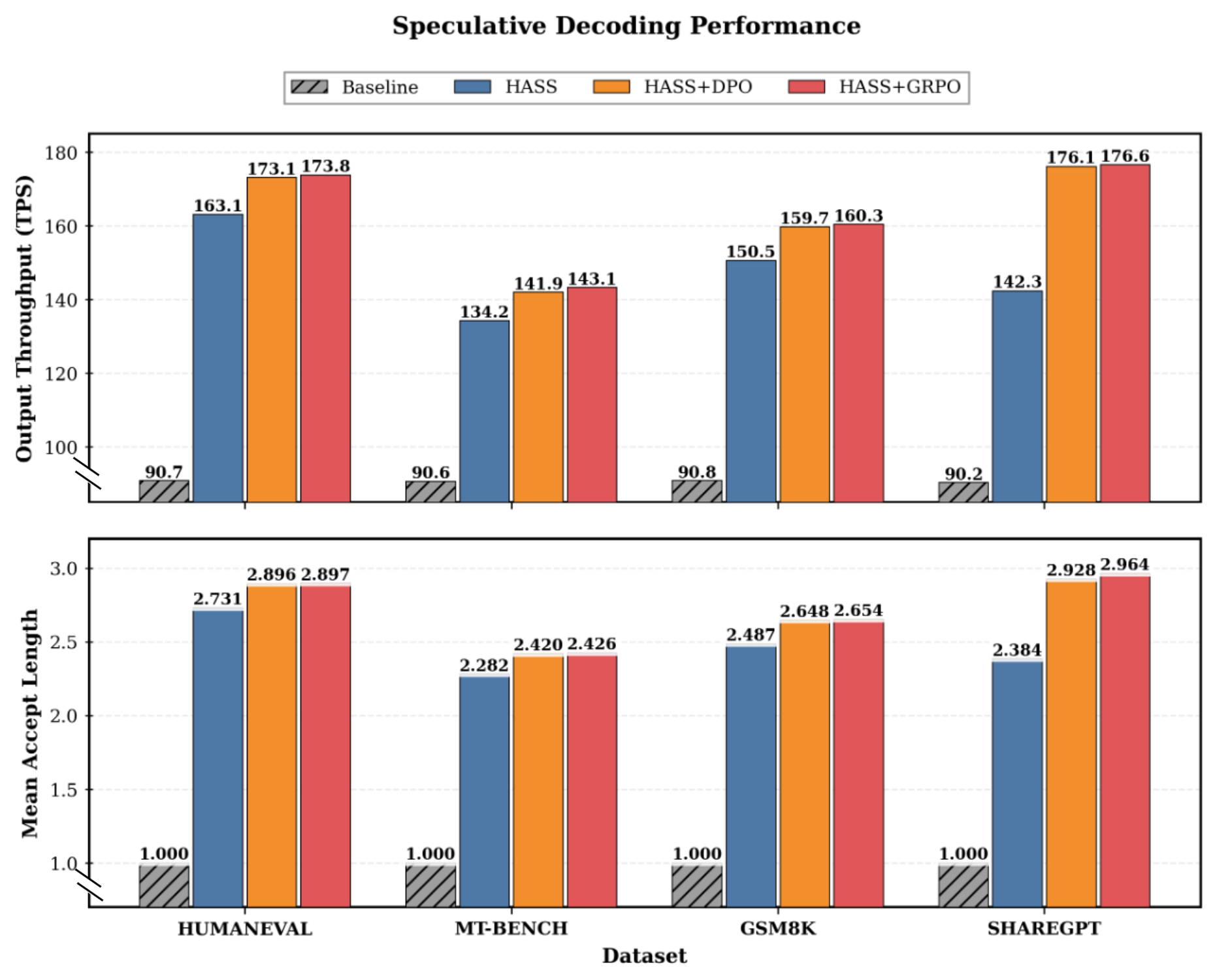}
  \caption{Output Throughput (TPS) and Mean Accept Length across model configurations. Evaluated on Llama-3.1-8B-Instruct with SGLang 5.5.0, the results demonstrate that SFT+DPO and SFT+GRPO consistently outperform the baseline SFT approach on both metrics.}
  \label{fig:perf_comparison}
\end{figure}

%% file: related-work.tex
\section{Related Work}
\label{sec:related_work}

\begin{figure}[h!]
\centering
\includegraphics[width=1\linewidth]{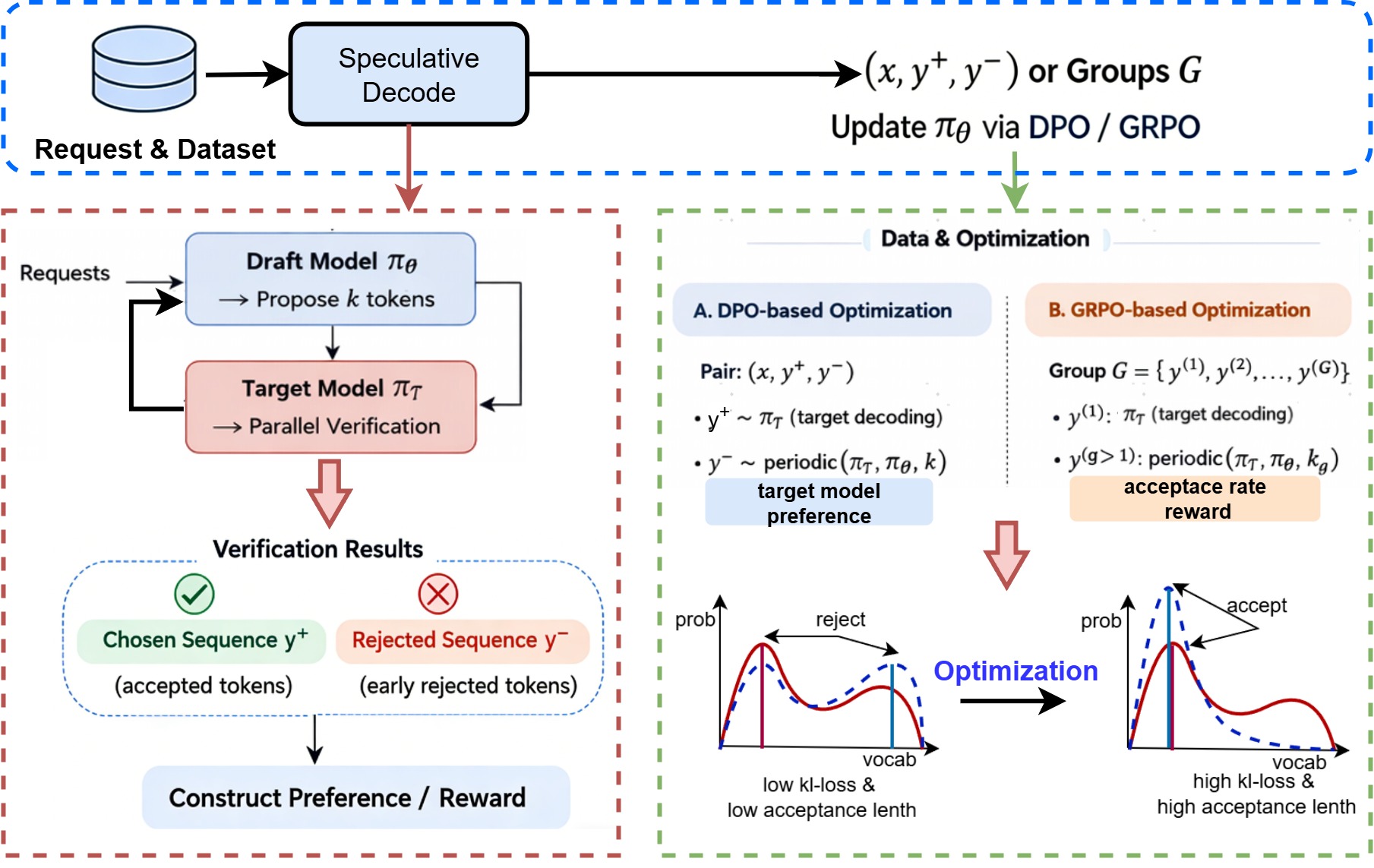}
\caption{Overview of the RCSD framework.
Given requests and dataset samples, a draft model $\pi_\theta$ first proposes $k$ tokens via speculative decoding, and a target model $\pi_T$ performs parallel verification.
The verification stage partitions outputs into accepted tokens (forming the chosen sequence $y^{+}$) and early-rejected tokens (forming the rejected sequence $y^{-}$), from which preference or reward signals are constructed.
The resulting pairs $(x, y^{+}, y^{-})$ or response groups $G$ are then used to update $\pi_\theta$ via DPO- or GRPO-based optimization.
In DPO-based optimization, target-model preferences between $y^{+}$ and $y^{-}$ guide parameter updates.
In GRPO-based optimization, grouped responses are optimized using an acceptance-rate reward.}
\label{fig:overall}
\end{figure}

\subsection{Speculative Decoding for Autoregressive Inference}
Speculative decoding has emerged as an effective paradigm for accelerating autoregressive inference in large language models by decoupling token proposal and verification.
The foundational work on speculative decoding introduces a draft--target--verification framework, where a lightweight draft model generates candidate tokens that are subsequently verified in parallel by a larger target model, enabling substantial throughput improvements when multiple tokens are accepted at once~\cite{specdecode}.

Subsequent works have focused on improving the efficiency and scalability of speculative decoding through architectural and algorithmic innovations.
Medusa proposes multi-head token prediction to increase the number of speculative candidates per verification step~\cite{medusa}.
Other approaches explore lookahead decoding and multi-token prediction strategies to further amortize verification costs~\cite{specinfer}.
More recently, EAGLE introduces hidden-state guidance to enhance draft token quality and improve acceptance rates~\cite{eagle}.
While these methods significantly advance speculative inference at the decoding level, they generally assume a pre-trained draft model and devote limited attention to the alignment between the draft model's training objective and the system-level acceptance efficiency.

\subsection{Draft Model Training via Knowledge Distillation}
The dominant approach to training draft models relies on knowledge distillation, where the draft model is supervised to mimic the output distribution of the target model.
Classic distillation methods minimize the forward Kullback--Leibler divergence between the draft and target distributions using soft labels~\cite{distilling}.
This paradigm has been widely adopted in speculative decoding frameworks due to its simplicity and stability~\cite{specdecode,eagle}.

However, minimizing forward KL divergence encourages mode-covering behavior, forcing the draft model to allocate capacity to low-probability regions of the target distribution.
For speculative decoding, this objective is misaligned with the system goal of maximizing the expected acceptance length, which primarily depends on accurately proposing high-probability token sequences.
As discussed in prior work on divergence optimization, KL-based objectives fundamentally favor coverage over decision optimality~\cite{reverse}.
Various heuristics, such as temperature scaling and truncated distillation, have been explored to mitigate this issue, but the core training objective remains distribution matching rather than acceptance-oriented optimization.

\subsection{Reinforcement Learning Methods}
Reinforcement learning has been extensively applied to language model alignment, most notably through RLHF pipelines that optimize policies using human feedback and KL-regularized objectives~\cite{RLHF}.
While effective, these methods require explicit reward modeling and incur significant training complexity.

Direct Preference Optimization (DPO) offers a simpler alternative by directly optimizing policies using pairwise preference data, eliminating the need for an explicit reward model~\cite{DPO}.
DPO and related preference-based methods have demonstrated strong performance in instruction tuning and alignment tasks.
In parallel, group-wise and relative policy optimization techniques, such as GRPO-style clipping and ranking-based learning, leverage relative comparisons across multiple candidates to stabilize learning and improve sample efficiency~\cite{GRPO}.

Existing applications of preference optimization primarily focus on improving response quality or alignment with human preferences.
In contrast, our work applies both pairwise and group-wise preference optimization directly to speculative decoding, where preferences and rewards are automatically derived from the target model and explicitly aligned with inference-time acceptance efficiency.

\subsection{Reinforcement Learning in Speculative Decoding}

Recent studies have explored integrating reinforcement learning (RL) into speculative inference systems. To date, these efforts have primarily focused on optimizing inference-time hyperparameters via RL, such as sampling tree depth, step size, and verification length~\cite{RL_spec_step}. While these methods demonstrate promising results in single-request settings, they encounter limitations in concurrent scenarios. Due to engineering constraints, requests within the same batch typically require uniform speculative decoding hyperparameters, restricting the ability to adapt parameters based on specific input characteristics. Consequently, per-instance hyperparameter tuning yields diminishing returns in batched inference, where simple adjustments based on batch size often suffice. Furthermore, hyperparameter in inference and prediction accuracy constitute two distinct determinants of inference speed within speculative decoding systems. Irrespective of the chosen inference hyperparameters, enhancing prediction accuracy invariably leads to improved inference throughput. Motivated by this observation, this work employs reinforcement learning to directly optimize the prediction accuracy of the draft model itself. Therefore, our approach is orthogonal to and independent of the aforementioned inference-time hyperparameter optimization methods.
Building on the preference optimization paradigms discussed above, we treat the target model as an implicit reward mechanism that simultaneously generates reference outputs, verifies speculative tokens, and provides preference signals. Unlike prior self-feedback methods that focus on output quality or safety, we leverage this intrinsic model feedback to optimize the acceptance length in speculative decoding \cref{fig:overall}. This approach eliminates the need for additional external reward models or human annotations, establishing a direct link between the training objective and the system-level performance metric. 

%% file: methodology.tex
\newcommand{\group}[1]{\mathbf{y}^{(#1)}}
\newcommand{\groupname}[1]{\text{#1}}

\section{Methodology}
In this section, we present two complementary optimization strategies for training the draft model in the Eagle\xspace\ speculative inference framework.
Both approaches leverage the target model ${\mathcal{M}_{\text{target}}}$ as an implicit reward signal to align the draft policy ${\pi_{\theta}}$ with the target model’s generation behavior.

We first introduce a pairwise preference-based optimization method based on Direct Preference Optimization (DPO).
This approach constructs binary preference signals by contrasting target-model generations with speculative generations produced by the draft model.
We then extend this formulation to a group-wise relative optimization strategy, referred to as Group-wise Relative Policy Optimization (GRPO), which exploits relative quality signals across multiple speculative generation paths.

Together, these methods enable efficient training of draft models that maximize token verification acceptance rates during speculative inference.

\subsection{Direct Preference Optimization for {Eagle\xspace} Training}
In the {Eagle\xspace} speculative inference framework, we optimize a policy draft-model ${\pi_{\theta}}$ via Direct Preference Optimization to align with a target model ${\mathcal{M}_{\text{target}}}$. The DPO dataset ${\mathcal{D}_{\text{dpo}}}$ consists of pairs $({\mathbf{y}^+}, {\mathbf{y}^-})$ where:
\begin{itemize}
    \item \textbf{Chosen sequences ${\mathbf{y}^+}$} are sampled directly from ${\mathcal{M}_{\text{target}}}$ using autoregressive decoding
    \item \textbf{Rejected sequences ${\mathbf{y}^-}$} are generated through periodic alternation: ${\mathcal{M}_{\text{target}}}$ generates 1 token, followed by ${\mathcal{M}_{\text{draft}}}$ generating blocks of size $k$ without verification
\end{itemize}
This creates an implicit preference ${\mathbf{y}^+} \succ {\mathbf{y}^-}$. Since the draft-model's input structure differs from the target model's, the sequence representation ${\mathbf{y}}$  incorporates not only tokens but also hidden states from ${\mathcal{M}_{\text{target}}}$.

\subsubsection{DPO Objective}
Let ${\pi_{\text{ref}}}$ denote a frozen reference draft-model (initialized from ${\pi_{\theta}}$'s pretrained weights). For sequence $\mathbf{y}$ given prompt $\mathbf{x}$, define the log-ratio:
\begin{equation}
    r_{\theta}(\mathbf{x}, \mathbf{y}) = \log {\pi_{\theta}}(\mathbf{y}_{>|\mathbf{x}|} \mid \mathbf{x}) - \log {\pi_{\text{ref}}}(\mathbf{y}_{>|\mathbf{x}|} \mid \mathbf{x})
\end{equation}
where $\mathbf{y}_{>|\mathbf{x}|}$ denotes tokens after the prompt. The {DPO\xspace} loss minimizes:
\begin{equation}
\label{eq:dpo_loss}
    {\mathcal{L}_{DPO\xspace}}(\theta) = -{E}_{(\mathbf{x}, {\mathbf{y}^+}, {\mathbf{y}^-}) \sim {\mathcal{D}_{\text{dpo}}}} \left[ 
        \log \sigma \left( \beta \left( 
            r_{\theta}(\mathbf{x}, {\mathbf{y}^+}) - r_{\theta}(\mathbf{x}, {\mathbf{y}^-}) 
        \right) \right)
    \right]
\end{equation}
with $\beta > 0$ controlling optimization strength and $\sigma(\cdot)$ the sigmoid function.

\begin{algorithm}[t]
\caption{DPO Training for Speculative Inference}
\label{alg:dpo_training}
\begin{algorithmic}[1]
\Require 
    Policy model ${\pi_{\theta}}$, frozen reference model ${\pi_{\text{ref}}}$, 
    preference dataset ${\mathcal{D}_{\text{dpo}}}$, temperature $\beta$
\Ensure Optimized policy parameters $\theta$
\State Initialize $\theta$ from pretrained weights
\While{training not converged}
    \State Sample batch $\mathcal{B} = \{(\mathbf{x}^{(i)}, {\mathbf{y}^+}^{(i)}, {\mathbf{y}^-}^{(i)})\}_{i=1}^B \sim {\mathcal{D}_{\text{dpo}}}$
    \State Compute log-probabilities for all sequences:
    \Statex $\quad$ $\log p_{\mathbf{y}^+}^{\pi_{\theta}} \gets \sum_{t > |\mathbf{x}|} \log {\pi_{\theta}}(y_t \mid \mathbf{x}_{<t})$ \Comment{Policy on chosen}
    \Statex $\quad$ $\log p_{\mathbf{y}^-}^{\pi_{\theta}} \gets \sum_{t > |\mathbf{x}|} \log {\pi_{\theta}}(z_t \mid \mathbf{x}_{<t})$ \Comment{Policy on rejected}
    \Statex $\quad$ $\log p_{\mathbf{y}^+}^{\pi_{\text{ref}}} \gets \sum_{t > |\mathbf{x}|} \log {\pi_{\text{ref}}}(y_t \mid \mathbf{x}_{<t})$ \Comment{Reference on chosen}
    \Statex $\quad$ $\log p_{\mathbf{y}^-}^{\pi_{\text{ref}}} \gets \sum_{t > |\mathbf{x}|} \log {\pi_{\text{ref}}}(z_t \mid \mathbf{x}_{<t})$ \Comment{Reference on rejected}
    \State Calculate preference ratios:
    \Statex $\quad$ $\Delta^{\pi_{\theta}} \gets \log p_{\mathbf{y}^+}^{\pi_{\theta}} - \log p_{\mathbf{y}^-}^{\pi_{\theta}}$
    \Statex $\quad$ $\Delta^{\pi_{\text{ref}}} \gets \text{detach}\left( \log p_{\mathbf{y}^+}^{\pi_{\text{ref}}} - \log p_{\mathbf{y}^-}^{\pi_{\text{ref}}} \right)$
    \State Compute loss: $\mathcal{L} \gets -\frac{1}{B} \sum_{i=1}^B \log \sigma \left( \beta (\Delta^{\pi_{\theta}}_i - \Delta^{\pi_{\text{ref}}}_i) \right)$
    \State Update $\theta \gets \theta - \eta \nabla_{\theta} \mathcal{L}$
\EndWhile
\end{algorithmic}
\end{algorithm}

\subsubsection{Training Protocol}
Algorithm \ref{alg:dpo_training} outlines the optimization procedure:
\begin{enumerate}
    \item \textbf{Sequence Generation:}
    \begin{itemize}
        \item ${\mathbf{y}^+}$ is sampled from ${\mathcal{M}_{\text{target}}}$ via standard autoregressive decoding
        \item ${\mathbf{y}^-}$ is generated through periodic alternation: for each position $t$,
        \begin{equation*}
            y_t = \begin{cases} 
                \text{sample from } {\mathcal{M}_{\text{draft}}}(\cdot \mid \mathbf{y}_{<t}) & \text{if } t \mod (k+1) \neq 0 \\
                \text{sample from } {\mathcal{M}_{\text{target}}}(\cdot \mid \mathbf{y}_{<t}) & \text{otherwise}
            \end{cases}
        \end{equation*}
    \end{itemize}
    
    \item \textbf{Log-Probability Calculation:} 
    Sequence log-probabilities are computed exclusively over post-prompt tokens ($t > |\mathbf{x}|$) with next-token prediction alignment.     Additionally, tokens in rejected sequences generated by ${\mathcal{M}_{\text{target}}}$ are masked to reduce penalty on tokens already aligned with ${\mathcal{M}_{\text{target}}}$. Given the length sensitivity of DPO, identical masking patterns (for both prompt exclusion and ${\mathcal{M}_{\text{target}}}$-generated tokens) are consistently applied across chosen and rejected sequences during probability calculations.
    
\end{enumerate}

\subsection{Group-wise Relative Policy Optimization for {Eagle\xspace} Training}
In the {Eagle\xspace} speculative inference framework, we optimize a policy draft-model ${\pi_{\theta}}$ via Group-wise Relative Policy Optimization to align with a target model ${\mathcal{M}_{\text{target}}}$. The {GRPO\xspace} dataset ${\mathcal{D}_{\text{grpo}}}$ consists of tuples $(\mathbf{x}, \{\group{g}\}_{g \in \mathcal{G}})$ where:
\begin{itemize}
    \item $\mathbf{x}$ is the input prompt
    \item $\mathcal{G}$ denotes the set of sequence groups (e.g., $\mathcal{G} = \{\text{chosen}, \text{spec\_step1}, \text{spec\_step2}, \text{spec\_step3}\}$)
    \item \textbf{Chosen group } $\group{\text{chosen}}$ is sampled directly from ${\mathcal{M}_{\text{target}}}$ using autoregressive decoding
    \item \textbf{Speculative groups } $\group{\text{spec\_step}k}$ are generated through periodic alternation: ${\mathcal{M}_{\text{target}}}$ generates 1 token, followed by ${\mathcal{M}_{\text{draft}}}$ generating blocks of size $k$ without verification
\end{itemize}
Sequence quality scores $s_g$ are computed by ${\mathcal{M}_{\text{target}}}$ for each group $g$. Unlike binary preferences, {GRPO\xspace} leverages relative quality signals across multiple generation paths to optimize the draft policy.

\subsubsection{{GRPO\xspace} Objective}
Let ${\pi_{\text{ref}}}$ denote a frozen reference draft-model (initialized from ${\pi_{\theta}}$'s pretrained weights). For sequence $\group{g}$ given prompt $\mathbf{x}$, define the probability ratio:
\begin{equation}
    r_{\theta}(\mathbf{x}, \group{g}) = \frac{{\pi_{\theta}}(\group{g}_{>|\mathbf{x}|} \mid \mathbf{x})}{{\pi_{\text{ref}}}(\group{g}_{>|\mathbf{x}|} \mid \mathbf{x})}
\end{equation}
where $\group{g}_{>|\mathbf{x}|}$ denotes tokens after the prompt. The normalized reward for group $g$ is:
\begin{equation}
    s_g = \mathcal{N}({\log {\mathcal{M}_{\text{target}}}(\group{g}_{>|\mathbf{x}|} \mid \mathbf{x})})+\mathcal{N}(Acc^{(g)}_{>|\mathbf{x}|})
\end{equation}
with $\mathcal{N}(\cdot)$ denotes Z-score normalization. The {GRPO\xspace} loss minimizes:

\begin{equation}
\label{eq:grpo_loss}
    {\mathcal{L}_{GRPO\xspace}}(\theta) = \frac{1}{|\mathcal{G}|}\sum_{g \in \mathcal{G}} {E}_{(\mathbf{x}, \group{g}) \sim {\mathcal{D}_{\text{grpo}}}} \left[ 
        -\min\left( 
            r_{\theta}(\mathbf{x}, \group{g}) \hat{s}_g, 
            \text{clip}_{1\pm\delta}\left(r_{\theta}(\mathbf{x}, \group{g})\right) \hat{s}_g 
        \right) + \beta \cdot \text{KL}_g
    \right]
\end{equation}

\begin{algorithm}[t]
\caption{{GRPO\xspace} Training for Speculative Inference}
\label{alg:grpo_training}
\begin{algorithmic}[1]
\Require Policy ${\pi_{\theta}}$, reference ${\pi_{\text{ref}}}$, dataset ${\mathcal{D}_{\text{grpo}}}$, $\beta$, $\delta$
\Ensure Optimized $\theta$
\State Initialize $\theta$ from pretrained weights
\While{not converged}
    \State Sample $\mathcal{B} = \{({\mathbf{x}}^{(i)}, \{{\group{g}}^{(i)}\}_{g \in \mathcal{G}})\}_{i=1}^B \sim {\mathcal{D}_{\text{grpo}}}$
    \State Compute normalized rewards $\forall g,i$:
    \Statex $\quad s_g^{(i)} \gets \mathcal{N}({\log {\mathcal{M}_{\text{target}}}(\group{g}_{>|\mathbf{x}|} \mid \mathbf{x})})+\mathcal{N}(Acc^{(g)}_{>|\mathbf{x}|})$
    \State $\mathcal{L} \gets 0$
    \For{each group $g \in \mathcal{G}$}
        \State $\Delta_g \gets \log {\pi_{\theta}}(\group{g}_{>|\mathbf{x}|} \mid \mathbf{x}) - \log {\pi_{\text{ref}}}(\group{g}_{>|\mathbf{x}|} \mid \mathbf{x})$ \Comment{Log-ratio}
        \State $r_g \gets \exp(\Delta_g)$, \quad $\tilde{r}_g \gets \text{clip}(r_g, 1\pm\delta)$
        \State $\text{KL}_g \gets \text{KL}[{\pi_{\theta}} \| {\pi_{\text{ref}}}]_{\group{g}}$ \Comment{Per-token KL}
        \State $\mathcal{L}_g \gets -{E}_{i} \left[ \min(r_g^{(i)} {s}_g^{(i)}, \tilde{r}_g^{(i)} {s}_g^{(i)}) \right] + \beta \cdot \text{KL}_g$
        \State $\mathcal{L} \gets \mathcal{L} + \mathcal{L}_g / |\mathcal{G}|$
    \EndFor
    \State Update $\theta \gets \theta - \eta \nabla_{\theta} \mathcal{L}$
\EndWhile
\end{algorithmic}
\end{algorithm}

\subsubsection{Training Protocol}
Algorithm \ref{alg:grpo_training} outlines the optimization procedure:
\begin{enumerate}
    \item \textbf{Group Configuration:}
    \begin{itemize}
        
        \item Quality scores $s_g$ are computed exclusively over post-prompt tokens ($t > |\mathbf{x}|$) using ${\mathcal{M}_{\text{target}}}$'s log-probabilities
        
    \end{itemize}
    
    \item \textbf{Sequence Processing:}
    \begin{itemize}
        \item Log-probabilities for ${\pi_{\theta}}$ and ${\pi_{\text{ref}}}$ are computed with next-token prediction alignment (hidden states $\mathbf{h}_t$ predict token $y_{t+1}$)
        
        \item Target-generated tokens in speculative sequences are masked during log-probability calculation to avoid over-penalizing aligned segments
        
        \item Identical masking patterns are applied consistently across all groups for fair reward comparison
    \end{itemize}
    
\end{enumerate}

%% file: experiments.tex
\section{Experiments}
\label{sec:experiments}

\subsection{Setup}

\paragraph{Models and datasets.}
We evaluate our methods on two model families: \textbf{Qwen3-1.7B} and \textbf{Llama-3.1-8B}, trained on 90\% of ShareGPT.
Evaluation uses GSM8K, HumanEval, MT-Bench, and the remaining 10\% of ShareGPT to cover both reasoning and open-ended instruction-following tasks.

\paragraph{Training variants.}
We compare: 
(1) HASS (baseline draft model), 
(2) HASS + DPO , 
(3) HASS + GRPO . 
Results are reported for speculative steps 3–6, i.e., drafting 3–6 candidate tokens per verification round, to examine performance across decoding configurations.

\subsection{Main Results}

To provide a holistic comparison, we compute the average acceptance length across the four evaluation datasets at each speculative step.

\paragraph{Qwen3-1.7B.}

\begin{table}[h]
\centering
\small
\setlength{\tabcolsep}{4pt}
\begin{tabular}{llcccc}
\toprule
\textbf{Dataset} & \textbf{Model} & \textbf{Step 3} & \textbf{Step 4} & \textbf{Step 5} & \textbf{Step 6} \\
\midrule
\multirow{3}{*}{HumanEval} 
& HASS & 2.322 & 2.446 & 2.510 & 2.533 \\
& HASS + DPO & 2.616 & 2.780 & 2.922 & 3.014 \\
& HASS + GRPO & \textbf{2.629} & \textbf{2.807} & \textbf{2.928} & \textbf{3.015} \\
\cmidrule{1-6}
\multirow{3}{*}{MT-Bench} 
& HASS & 1.986 & 2.047 & 2.072 & 2.092 \\
& HASS + DPO & 2.333 & 2.442 & 2.537 & 2.571 \\
& HASS + GRPO & \textbf{2.346} & \textbf{2.467} & \textbf{2.549} & \textbf{2.601} \\
\cmidrule{1-6}
\multirow{3}{*}{GSM8K} 
& HASS & 2.102 & 2.155 & 2.176 & 2.176 \\
& HASS + DPO & 2.746 & 3.001 & 3.121 & 3.212 \\
& HASS + GRPO & \textbf{2.754} & \textbf{3.002} & \textbf{3.161} & \textbf{3.246} \\
\cmidrule{1-6}
\multirow{3}{*}{ShareGPT} 
& HASS & 2.050 & 2.166 & 2.180 & 2.220 \\
& HASS + DPO & \textbf{2.581} & 2.880 & 3.042 & 3.089 \\
& HASS + GRPO & 2.563 & \textbf{2.882} & \textbf{3.051} & \textbf{3.136} \\
\bottomrule
\end{tabular}
\caption{Qwen3-1.7B: results at Steps 3--6. Baseline HASS, HASS+DPO, and HASS+GRPO are compared. Higher values indicate better acceptance length. HASS+GRPO generally achieves better performance.}
\label{tab:qwen17b}
\end{table}

Table~\ref{tab:qwen17b} shows the results for Qwen3-1.7B. RL significantly improves over the baseline across all datasets and speculative steps. Averaged over the four datasets, acceptance length at step 6 reaches 2.26, 2.97, and 3.00 for HASS, HASS+DPO, and HASS+GRPO, respectively. GRPO generally achieves the best performance, with consistent gains in GSM8K and MT-Bench.

\paragraph{Llama-3.1-8B.}

\begin{table}[h]
\centering
\small
\setlength{\tabcolsep}{4pt}
\begin{tabular}{llcccc}
\toprule
\textbf{Dataset} & \textbf{Model} & \textbf{Step 3} & \textbf{Step 4} & \textbf{Step 5} & \textbf{Step 6} \\
\midrule
\multirow{3}{*}{HumanEval} 
& HASS & 2.731 & 2.950 & 3.103 & 3.202 \\
& HASS + DPO & 2.896 & 3.185 & 3.391 & 3.540 \\
& HASS + GRPO & \textbf{2.897} & \textbf{3.189} & \textbf{3.402} & \textbf{3.558} \\
\cmidrule{1-6}
\multirow{3}{*}{MT-Bench} 
& HASS & 2.282 & 2.397 & 2.438 & 2.497 \\
& HASS + DPO & 2.420 & 2.595 & 2.674 & 2.731 \\
& HASS + GRPO & \textbf{2.426} & \textbf{2.596} & \textbf{2.692} & \textbf{2.736} \\
\cmidrule{1-6}
\multirow{3}{*}{GSM8K} 
& HASS & 2.487 & 2.640 & 2.712 & 2.756 \\
& HASS + DPO & 2.648 & 2.836 & 2.945 & 3.008 \\
& HASS + GRPO & \textbf{2.654} & \textbf{2.864} & \textbf{2.958} & \textbf{3.025} \\
\cmidrule{1-6}
\multirow{3}{*}{ShareGPT} 
& HASS & 2.384 & 2.524 & 2.639 & 2.672 \\
& HASS + DPO & 2.928 & 3.283 & 3.571 & 3.779 \\
& HASS + GRPO & \textbf{2.964} & \textbf{3.310} & \textbf{3.588} & \textbf{3.828} \\
\bottomrule
\end{tabular}
\caption{Llama-3.1-8B: results at Steps 3--6. HASS+GRPO generally achieves better performance.}
\label{tab:llama8b}
\end{table}

Table~\ref{tab:llama8b} shows similar trends for Llama-3.1-8B. 
Across most datasets and steps, GRPO achieves the highest or tied-best performance, with particularly strong improvements on ShareGPT. These results indicate that preference optimization substantially improves acceptance length, while GRPO consistently provides the best overall performance.

\subsection{Analysis}
Across both model families and all speculative steps, preference-based optimization consistently improves acceptance length over the HASS distillation baseline. GRPO tends to outperform DPO, highlighting the advantage of leveraging multiple speculative paths. Moreover, the relative gain grows as more tokens are drafted per step (e.g., on Qwen3-1.7B + GSM8K, from 31\% at step 3 to 49\% at step 6), suggesting that better draft quality compounds over longer speculation horizons.

%% file: conclusion.tex
\section{Conclusion}
\label{sec:conclusion}

We propose \textbf{RCSD} (Reinforcement and Continual Learning for Speculative Decoding), a training framework that fundamentally reorients draft model optimization from distribution matching to acceptance-length maximization—the true system-level objective of speculative decoding. By leveraging the intrinsic feedback loop inherent in speculative inference, RCSD constructs implicit preference signals from naturally occurring accepted and rejected token sequences without requiring external reward models or human annotations. Through two complementary formulations—pairwise preference optimization via DPO and group-wise relative policy optimization via GRPO—RCSD trains draft models to adopt mode-seeking behavior that prioritizes high-probability token sequences aligned with the target model's verification decisions. Crucially, our masked loss mechanism ensures training signal purity by excluding target-generated tokens from gradient updates.

Extensive experiments on Qwen3-1.7B and Llama-3.1-8B across diverse benchmarks (ShareGPT, GSM8K, HumanEval, and MT-Bench) demonstrate RCSD's effectiveness. Our method improves average acceptance length by \textbf{35--40\%} over conventional distillation baselines (e.g., HASS) and achieves \textbf{1.2--1.35$\times$} end-to-end inference speedup on practical workloads while preserving generation quality. Notably, GRPO consistently accelerates early-stage training convergence by leveraging relative quality signals across multiple speculative paths, highlighting the advantage of group-wise optimization over binary preferences.